\documentclass{article}

\usepackage{arxiv}
\usepackage[utf8]{inputenc} 
\usepackage[T1]{fontenc}    
\usepackage{hyperref}       
\usepackage{url}            
\usepackage{booktabs}       
\usepackage{amsfonts}       
\usepackage{nicefrac}       
\usepackage{microtype}      
\usepackage{lipsum}
\usepackage{graphicx}

\usepackage{bm}
\usepackage{amsmath}
\usepackage{siunitx}
\usepackage{titlesec}
\usepackage{footnote}
\usepackage{booktabs}
\makesavenoteenv{tabular}
\makesavenoteenv{table}
\usepackage{url}  
\usepackage[boldmath]{numprint}

\title{Who Needs Words? Lexicon-Free Speech Recognition}

\author{
  Tatiana Likhomanenko \\
  Facebook AI Research\\
  Menlo Park, USA \\
  \texttt{antares@fb.com} \\
  \And
  Gabriel Synnaeve \\
  Facebook AI Research\\
  New York, USA \\
  \texttt{gab@fb.com} \\
  \And
  Ronan Collobert \\
  Facebook AI Research\\
  Menlo Park, USA \\
  \texttt{locronan@fb.com} \\
}

\begin{document}
\maketitle

\begin{abstract}
Lexicon-free speech recognition naturally deals with the problem of out-of-vocabulary (OOV) words. In this paper, we show that character-based language models (LM) can perform as well as word-based LMs for speech recognition, in word error rates (WER), even without restricting the decoding to a lexicon. We study character-based LMs and show that convolutional LMs can effectively leverage large (character) contexts, which is key for good speech recognition performance downstream. We specifically show that the lexicon-free decoding performance (WER) on utterances with OOV words using character-based LMs is better than lexicon-based decoding, both with character or word-based LMs.
\end{abstract}

\keywords{speech recognition, beam-search decoder, out-of-vocabulary words, lexicon-free}

\section{Introduction}
\label{S:Intro}

Character-based models permeated text classification \cite{zhang2015character}, language modeling \cite{potamianos1998study,carpenter2005scaling,kim2016character}, machine translation \cite{vilar2017canwe,nakov2012combining,sennrich2015neural,ling2015character,costa2016character,chung2016character}, and automatic speech recognition (ASR) \cite{graves2014towards,chan2016listen,bahdanau2016end}. However, most competitive ASR systems, character based or not, use a beam search decoder constrained on a word-level language model and lexicon \cite{amodei2016deep,liptchinsky2017letterbased,zhang2017towards,zeghidour2018fully}. 
In recent works \cite{zeyer2018improved,chiu2018state}, authors achieved competitive results with acoustic models (AM) and LMs operating on word pieces, and a lexicon-free decoder.
To the best of our knowledge, the first ASR system to achieve competitive results with a character-based LM and without a lexicon (on Switchboard and WSJ) was \cite{hadian2018end}, that our lexicon-free character-based ConvLM surpasses on WSJ (see Table~\ref{T:Decoder:WSJ}).

The main advantage of a lexicon-free approach is that it allows the decoder to handle out-of-vocabulary (OOV) words: the decoder and the language model are responsible not only for scoring words but usually also restrict the vocabulary.
Drawbacks sometimes include system complexity and most often poorer performance than in the lexicon based case. The first lexicon-free beam-search decoder aiming at dealing with OOV was benchmarked on Switchboard \cite{maas2015lexicon}, although with a significantly worse word error rate (WER) than lexicon-based systems. 
Other recent works in this direction include \cite{hwang2016character,zenkel2017comparison} on the English and \cite{ahmed2016lexicon,smit2017character} on the Arabic and Finnish languages.

Here, we study a simple end-to-end ASR system combining a character level acoustic model with a character level language model through beam search. We show that it can yield competitive word error rates on the WSJ and Librispeech corporas, even without a lexicon. Finally, our model shows significant word error rates improvement on utterances that include out-of-vocabulary words.

\section{Setup}
\paragraph{Acoustic model (AM)} 
We consider in this paper 1D gated convolutional neural networks~\cite{dauphin2016language,liptchinsky2017letterbased}, trained to map speech features (log-mel filterbanks) to their corresponding letter transcription. The training criterion is the auto segmentation criterion (ASG)~\cite{collobert2016wav2letter}. The token set contains 31 graphemes: the standard English alphabet, the apostrophe and period, two repetition characters (e.g. the word \texttt{ann} is transcribed as \texttt{an1}), and a silence token (\texttt{|}) used as word boundary.

\paragraph{Language model (LM)}
Our language models are character-based. We evaluated $n$-gram language models, as well as gated convolutional language models (ConvLMs)~\cite{dauphin2016language}, and show that with enough context these language models can match (in perplexity) their word-based counterparts. The LM training data was pre-processed to be consistent with the AM training data: the silence character (\texttt{|}) defines word boundaries, and repetition symbols are used when letter repetitions occur.

\paragraph{Beam-search decoder}
We extended the beam-search decoder from \cite{liptchinsky2017letterbased} to support character-level language models. Given
a word transcription $y$, we denote $\text{AM}(y)$ the corresponding acoustic score and $P_{\text{LM}}(y)$ the corresponding LM likelihood.
The beam-search decoder generates transcriptions by finding the argmax of the following score \cite{collobert2016wav2letter}:
\begin{equation}
\text{AM}(y) + \alpha \log{P_{\text{LM}}(y)} + \beta|y| + \gamma \sum_{i=1}^T [\pi_i=`|`],
\end{equation}
where $\pi = \pi_1, ..., \pi_T$ is the sequence of letters corresponding to the transcription~$y$. The hyper-parameters $\alpha$, $\beta$ and $\gamma$ weight the language model, word penalty and silence penalty, respectively. The decoder has two additional parameters: (i)~the beam size and (ii)~a beam threshold, controlling which hypothesis can make it to the beam.

\paragraph{Experiments}

We experiment with the Wall Street Journal (WSJ) dataset~\cite{garofolo1993csr,linguistic1994csr,woodland1994large} (about 81 hours of transcribed audio data) and the Librispeech dataset~\cite{panayotov2015librispeech} (1000 hours with clean and noisy speech). 
  
\section{Language Model Experiments}
\label{sec:explm}
We consider word-level $n$-gram and ConvLM-based language models as baseline, and compare them in word perplexity with their character-level counterpart.

\paragraph{Data preparation} 
Language models for both WSJ and Librispeech are trained with the corresponding language model data available for these datasets. For word-level model training, we keep all words (162K) for WSJ and use only the most frequent 200K (out of 900K) words for Librispeech (words appearing less than 10 times are dropped). Words outside this scope are replaced by \texttt{unknown}.

\paragraph{n-gram LMs} All models were trained with KenLM~\cite{heafield2013scalable}. For both Librispeech and WSJ, we trained $4$-gram word-level language models as a baseline. For character-level language models, we study how the context width impacts perplexity, training $n$-grams ranging from $5$ to $20$. For large values of $n$, we pruned the models by thresholding rarely-occurring $n$-grams: 6,7,8-grams appearing once, 9-grams appearing once or twice, and all $n$-grams for $n \geq 10$ appearing $\leq3$ times were dropped.

\paragraph{ConvLMs}
As a baseline for ConvLMs, we use the `GCNN-14B` word-level LM architecture from~\cite{dauphin2016language}, which achieved competitive results on several language model benchmarks.
This network contains 14 convolutional-residual blocks with a growing number of channels and gated linear units as activation functions, resulting in 318M parameters and an effective receptive field of 57 tokens. An {\it adaptive softmax} \cite{grave2017efficient} over words follows the convolutional layers.

For character-level LMs, we consider both the `GCNN-14B` architecture and a deeper variant (20 convolutional layers) dubbed `GCNN-20B`, with a larger receptive field of 81 tokens. For both configurations a {\it softmax} (over letters) follows the last convolutional layer. The resulting number of parameters was 163M for `GCNN-14B` and 224M for `GCNN-20B`. Dropout is used at each convolutional and linear layer: with probabilities 0.2 for WSJ and 0.1 for Librispeech.

ConvLMs were trained with the {\it fairseq} toolkit\footnote{\url{https://github.com/pytorch/fairseq}}~\cite{gehring2017convs2s}, using Nesterov accelerated gradient descent~\cite{nesterov1983method} with fixed learning rate. 
Gradient clipping and weight normalization are used following~\cite{dauphin2016language}.

\begin{table}[t!]
\caption{Word perplexity on the validation set of WSJ. For character-level LMs, we display lower and upper perplexity bounds. For models marked with \texttt{*}, pruning is applied during training. The receptive field is in characters.}
\label{T:LM:WSJ}
\centering
\sisetup{round-mode=places}
\begin{tabular}{l c c c}
\toprule
Language Model & Size & Rcp. field (char.) & nov93dev \\
\midrule
word 4-gram & 878 M & 32 & 156 \\
char 5-gram & 3.3 M & 5 & (714, 1285) \\
char 10-gram & 447 M & 10 & (211, 243) \\
char 15-gram$^\texttt{*}$ & 546 M & 15 & (185, 205)  \\
char 15-gram & 3.5 G & 15 & (186, 203) \\ 
char 20-gram$^\texttt{*}$ & 836 M & 20 & (177, 196) \\ 
char 20-gram & 9.7 G & 20 & (180, 196) \\ 
\midrule
word GCNN-14B & 1.1 G & 450 & 80 \\
char GCNN-14B & 936 M & 57 & (79, 95) \\
char GCNN-20B & 1.3 G & 81 & (76, 90) \\
\bottomrule
\end{tabular}
\end{table}

\begin{table}[t!]
\caption{Word perplexity on the validation sets of Librispeech, shown as lower and upper bounds for character-level LMs. For models marked with \texttt{*}, pruning is applied during training. The average receptive fields are 31 characters and 439 characters for the word 4-gram and word GCNN-14B LMs, respectively.}
\label{T:LM:Librispeech}
\centering
\sisetup{round-mode=places}
\begin{tabular}{l c c c}
\toprule
Language Model & Size & $\text{dev-clean}$ & $\text{dev-other}$ \\
\midrule
word 4-gram$^\texttt{*}$ & 13 G & 148 & 137 \\
char 5-gram & 7.7 M & (597, 1000) & (515, 869) \\
char 10-gram & 2.5 G & (198, 230) & (180, 210) \\
char 15-gram$^\texttt{*}$ & 6.5 G & (165, 180) & (151, 165) \\
char 17-gram$^\texttt{*}$ & 9.5 G & (163, 178) &  (148, 162) \\
char 20-gram$^\texttt{*}$ & 13 G & (162, 177) &  (147, 161) \\
\midrule
word GCNN-14B$^\texttt{*}$ & 1.8 G & 57  & 58 \\
char GCNN-14B & 936 M & (72, 88) & (70, 84) \\
char GCNN-20B & 1.3 G & (65, 76) & (64, 75) \\
\bottomrule
\end{tabular}
\end{table}

\paragraph{Word-level perplexity for character-level LMs}
To compare word and character-level LMs, we estimate a word-level perplexity\footnote{Defined as $\left(\prod_{i=1}^N p(\text{word}_i)\right)^{-\frac{1}{N}}$, $N$ is a number of words in data.} for character-level LMs. The
word probability can be estimated with:
\begin{equation}
\label{form:word}
P(\text{word}|\mathcal{C}) = P(l_1|\mathcal{C})\prod_{i=2}^n P(l_i|\mathcal{C}\text{ }l_{1}\dots l_{i-1}),
\end{equation}
where $l_1, ..., l_n$ are letters in a word and the last letter $l_n$ is a silence symbol $`|`$ with which the word finishes, $\mathcal{C}$ --- the previous context. However, this approach does not take into account that word-level LMs
are constrained to a fixed-size lexicon, while character-based LMs have virtually an infinite vocabulary. We thus re-normalize~(\ref{form:word}), taking into account only words from the word-level LM vocabulary $\mathcal{V}$:
\begin{equation}
\label{form:denom}
P_\mathcal{V}(\text{word}|\mathcal{C}) = \frac{P(\text{word}|\mathcal{C})} {\sum_{\text{word}_i\in \mathcal{V}}P(\text{word}_i|\mathcal{C})}.
\end{equation}
For large vocabulary $\mathcal{V}$ the denominator in (\ref{form:denom}) is computationally expensive. 
The probability (\ref{form:word}) can be used as an upper bound of (\ref{form:denom}) (see, for example, \cite{krause2016multiplicative}) while the lower bound can be obtained by taking in the denominator of (\ref{form:denom}) the sum over most probable (by word-level LM) words which cover 95\% of word-level LM distribution.

We then exclude from the perplexity computation words which are not presented in the word-level LM vocabulary $\mathcal{V}$ ($n$-gram and ConvLM models have the same $\mathcal{V}$).\footnote{Only about 20 word occurrences for {\it nov93dev} in WSJ, and around 200 (300) for {\it clean} ({\it other}) in Librispeech.}

\paragraph{Results}
The comparison of different language models is presented in Tables~\ref{T:LM:WSJ} and~\ref{T:LM:Librispeech} for WSJ and Librispeech, respectively. 
It can be seen that (as expected) increasing the context decreases perplexity for both $n$-gram models and ConvLMs. With $n$-grams, pruning is critical to avoid overfitting.
On both benchmarks character-level language models already have similar performance for $n$-grams with $n\geq15$, and are clearly outperformed by ConvLMs. With enough context, character-level LMs appear to be in the same ballpark as word-level LMs.

\section{ASR Experiments}
\label{sec:expasr}
In this section, we decode the output of a single acoustic model trained on WSJ or Librispeech, through a beam-search procedure constrained by the LMs trained in Section~\ref{sec:explm}. Both AM training and decoding were performed with the {\it wav2letter++} open source library\footnote{\url{https://github.com/facebookresearch/wav2letter}}~\cite{pratap2018w2l}. The decoder was adapted to support character-level LMs, as well as lexicon-free decoding, alleviating the need for a word lexicon while decoding.

\begin{table}[h!]
\caption{Word and character error rates (\%) on Librispeech data.}
\label{T:Decoder:Librispeech}
\centering
\sisetup{round-mode=places}
\begin{tabular}{l c S[round-precision=1] S[round-precision=1] S[round-precision=1] S[round-precision=1] S[round-precision=1] S[round-precision=1] S[round-precision=1] S[round-precision=1]}
\toprule
Language Model & Lexicon &  \multicolumn{2}{c}{dev-clean} & \multicolumn{2}{c}{dev-other} & \multicolumn{2}{c}{test-clean} & \multicolumn{2}{c}{test-other} \\
 & & \text{WER} & \text{CER} & \text{WER} & \text{CER} & \text{WER} & \text{CER}  & \text{WER} & \text{CER} \\
\midrule
CAPIO (Ensemble) \cite{han2017capio} & yes &  2.68 & & 7.56 & & 3.19 & & 7.64 & \\ 
CAPIO (Single)\footnotemark \cite{han2017capio}  & yes & 3.02 & & 8.28 & & 3.51 & & 8.58 & \\ 
Learnable front-end \cite{zeghidour2018fully} & yes & 3.16 & & 10.05 & & 3.44 & & 11.24 & \\ 
DeepSpeech2\footnotemark \cite{amodei2016deep} & yes &  & & & & 5.15 & & 12.73 & \\ 
Sequence-to-sequence \cite{zeyer2018improved} & yes & 3.54 & & 11.52 & & 3.82 & & 12.76 & \\ 
\midrule
word 4-gram$^\texttt{*}$ &  yes & 4.63402	& 1.80027	& 13.2979	& 6.80727  &  5.0194 & 1.84421 & 14.3801 & 7.20492\\
char 15-gram$^\texttt{*}$ & yes & 4.79946 & 1.85713 & 14.1085 & 7.125 & 5.20199	& 1.90459	& 14.9285	& 7.43626  \\
char 20-gram$^\texttt{*}$ & yes & 4.80865 & 1.86025 & 13.8592 & 6.99302 & 5.12211	& 1.86623
 & 14.7699 & 7.31381 \\ 
char 15-gram$^\texttt{*}$ & no & 4.86195 & 1.87585	& 14.0908 & 7.07236 & 5.10499 & 1.87618 & 14.8559 & 7.37357 \\
char 20-gram$^\texttt{*}$ & no & 4.82151 & 1.86233 & 13.8514 & 6.93887 & 4.9281 & 1.81224 & 14.747 & 7.21885 \\ 
\midrule
word GCNN-14B$^\texttt{*}$ & yes & 3.1451049972427483 & 1.2542612622375684 &  10.336024133626443 & 5.431453210027412 & 3.368459275524954 & 1.2684254023549888  & 11.069304306975145 & 5.670593850922796 \\
char GCNN-20B & yes & 3.38958 & 1.36347 & 10.7188 & 5.72023 & 3.72223 & 1.39701 & 11.6978 & 6.07498 \\
char GCNN-20B & no & 3.39326 & 1.35757 & 10.7639 & 5.71271 & 3.53774 & 1.34018 & 11.736 & 6.08965 \\
\bottomrule
\end{tabular}
\end{table}
\addtocounter{footnote}{-2} 
\stepcounter{footnote}\footnotetext{Speaker adaptation; pronunciation lexicon }
\stepcounter{footnote}\footnotetext{\label{noteDS2}12k hours AM train set and common crawl LM }

\begin{table}[h!]
\caption{Word and character error rates (\%) on WSJ data.}
\label{T:Decoder:WSJ}
\centering
\sisetup{round-mode=places}
\begin{tabular}{l c S[round-precision=1]
S[round-precision=1] S[round-precision=1] S[round-precision=1]}
\toprule
Language Model & Lexicon &  \multicolumn{2}{c}{nov93dev} & \multicolumn{2}{c}{nov92} \\
  &  & \text{WER} & \text{CER} & \text{WER} & \text{CER} \\
\midrule
Regular LF-MMI \cite{hadian2018end} & yes &  & & 2.8 &  \\
DeepSpeech2$^\text{\ref{noteDS2}}$ \cite{amodei2016deep} & yes &  &  & 3.1 & \\ 
CNN-BLSTM-HMM\footnote{Speaker adaptation; 3k acoustic states} \cite{chan2015deep} & yes & 6.6 & & 3.5 & \\ 
Learnable front-end \cite{zeghidour2018fully} & yes & 6.8 & & 3.5 & \\ 
EE-LF-MMI (word LM)\footnote{\label{noteLF}Data augmentation; $n$-gram LM} \cite{hadian2018end} & yes &  & & 4.1 &  \\
EE-LF-MMI (char LM)$^\text{\ref{noteLF}}$ \cite{hadian2018end} & no &  & & 5.4 &  \\
\midrule
word 4-gram &  yes & 8.51348 & 3.44436 & 5.52897 & 2.21141 \\
char 15-gram$^\texttt{*}$ & yes & 9.44863 & 3.73105 & 6.20237 & 2.31957 \\
char 20-gram$^\texttt{*}$ & yes & 9.25431 & 3.63824 & 5.91884 & 2.18436 \\
char 15-gram$^\texttt{*}$ & no & 9.5215 & 3.7558 & 6.16693 & 2.30755 \\
char 20-gram$^\texttt{*}$ & no & 9.3879 & 3.71043 & 6.09605 & 2.2715 \\
\midrule
word GCNN-14B & yes & 6.50959 & 2.6998 & 4.25306 & 1.77573 \\
char GCNN-20B & yes & 6.44887 & 2.8153 & 3.63282 & 1.52335 \\
char GCNN-20B & no & 6.44887 & 2.71424 & 3.6151 & 1.51133\\
\bottomrule
\end{tabular}
\end{table}

\paragraph{Data preparation}
For WSJ, we consider the standard subsets {\it si284}, {\it nov93dev} and {\it nov92} for training, validation and test, respectively. 
For Librispeech, all the available training data was used for training. 
Validation and test were achieved according to the available two configurations ({\it clean} for clean speech and {\it other} for "noisy" speech). 
All hyper-parameter tuning was performed on validation sets, and only final performance was evaluated on the test sets.
We kept the original 16kHz sampling rate and computed log-mel filterbanks with 40 (for Librispeech) or 80 (for WSJ) coefficients for a 25ms sliding window, strided by 10ms. 
All features are normalized to have zero mean and unit variance per input sequence before feeding into the neural network.
No data augmentation or speaker adaptation was performed. 

\paragraph{Acoustic model training} 
Models are trained with stochastic gradient descent (SGD), gradient clipping~\cite{pascanu2013difficulty} and weight normalization~\cite{salimans2016weight}. 
We followed  \cite{liptchinsky2017letterbased} for the architecture choices, picking the "high dropout" model with 19 convolutional layers for Librispeech, and the lighter version with 17 layers for WSJ. 
Batch size was set to 4 and 16, for Librispeech, and WSJ respectively.

\paragraph{Tuning the beam-search decoder}
Hyper-parameters of the decoder were selected via a random search. 
A large fixed beam size and beam threshold were set before running a random search.
The LM weight $\alpha$ was randomly sampled from the interval $(0, 5)$, the word $\beta$ and silence $\gamma$ penalties were sampled from the interval $(-5, 5)$ for both language model types. 
For each configuration and dataset up to 100 attempts of random search were run. 
Hyper-parameters that lead to the best WER were chosen for the final evaluation on the test sets. 

\begin{figure}[b!]
    \centering
    \includegraphics[width=0.7\textwidth]{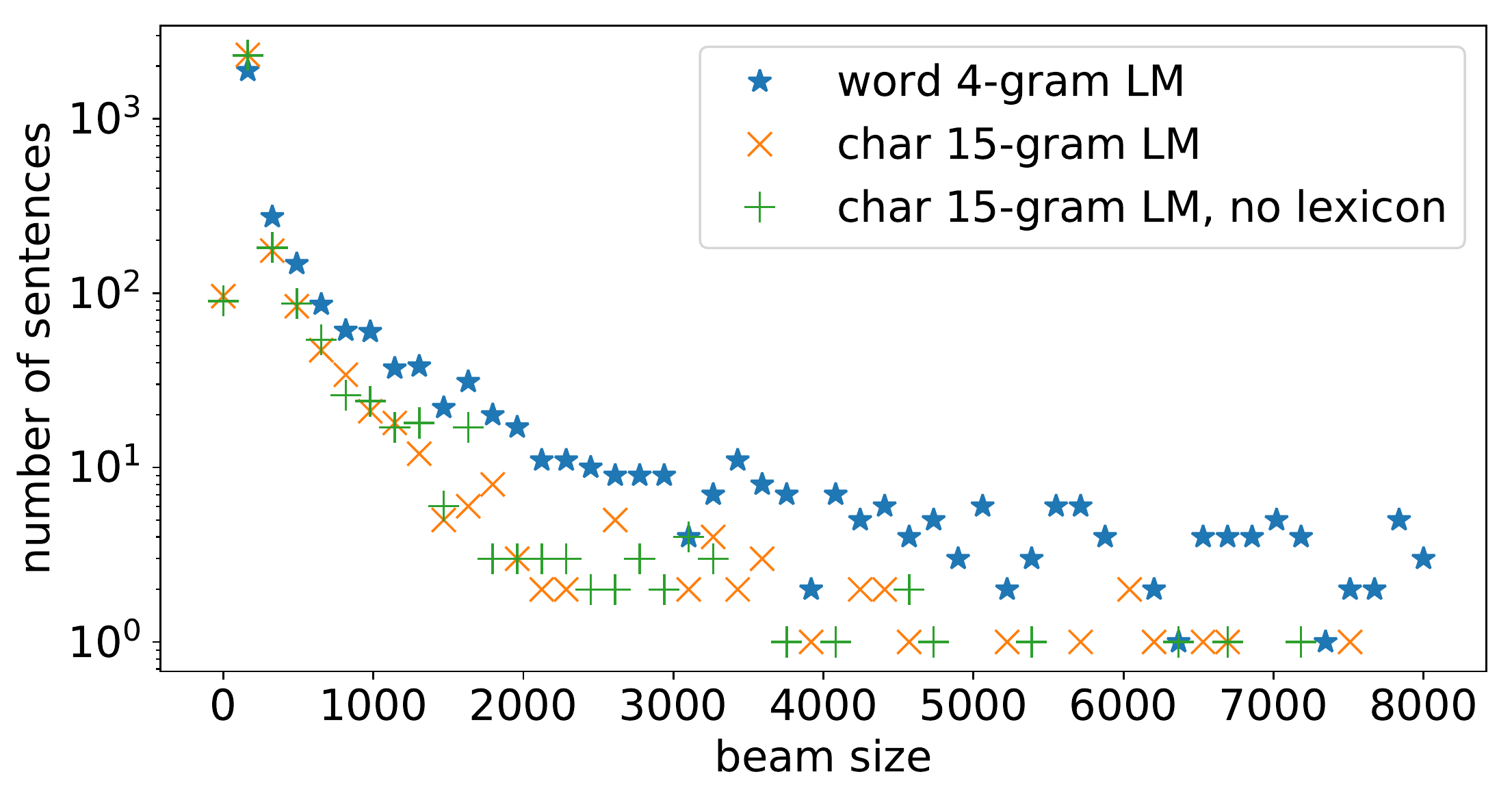}
    \caption{Effective beam size for the beam-search decoder (8000 beam size is set) with different LMs on `other` part of Librispeech validation set.}
    \label{Fig:beam_pdf}
\end{figure}

\paragraph{Results}
Models are evaluated in Word Error Rate (WER) and Character Error Rate (CER), as reported in Table~\ref{T:Decoder:WSJ} for WSJ and in Table~\ref{T:Decoder:Librispeech} for Librispeech.
The ASR system based on either $n$-gram or ConvLM character-level language model achieves performance similar to its word-level language model configuration both on WSJ and Librispeech. Furthermore, the lexicon-free ASR systems, where the beam-search is not conditioned by a word-level lexicon vocabulary, are very close to their lexicon-based counterparts for both types of language models, $n$-gram and ConvLM.
On the WSJ test set (see Table~\ref{T:Decoder:WSJ}), decoding with the character-level ConvLM (even in a lexicon-free setup) leads to better performance than with word-level and achieves state-of-the-art results.

\begin{table}[t!]
\caption{Word and character error rates (\%) for Librispeech: `in vocabulary` (IV) utterances and `out-of-vocabulary` (OOV) utterances. Proportions for test-clean data:~IV~---~~93.3\% utterances, OOV --- ~6.7\% utterances (176  among 2620). Proportions for test-other data:~IV~---~91.6\% utterances, OOV~--- ~8.4\% utterances (247  among 2939). }
\label{T:Decoder:Librispeech_unk}
\centering
\sisetup{round-mode=places}
\begin{tabular}{l c S[round-precision=1] S[round-precision=1] S[round-precision=1] S[round-precision=1] S[round-precision=1] S[round-precision=1] S[round-precision=1] S[round-precision=1]}
\toprule
Language Model & Lexicon & 
\multicolumn{2}{c}{test-clean IV} & 
\multicolumn{2}{c}{test-other IV} &
\multicolumn{2}{c}{test-clean OOV} & \multicolumn{2}{c}{test-other OOV} \\
 & & \text{WER} & \text{CER}  & \text{WER} & \text{CER} & \text{WER} & \text{CER}  & \text{WER} & \text{CER} \\
\midrule
word 4-gram$^\texttt{*}$ &  yes & 4.43593 & 1.64884 & 13.4632 & 6.75731 & 11.1647 & 3.84477 & 22.1598 & 10.8413 \\
char 15-gram$^\texttt{*}$ & yes & 4.61087 & 1.71629 & 14.0056 & 7.00724 & 11.4279 & 3.83279 & 22.7577 & 10.9216\\
char 20-gram$^\texttt{*}$ & yes & 4.49008 & 1.68276 & 13.8711 & 6.88125 & 11.7789 & 3.74496 & 22.3954 & 10.8279 \\
char 15-gram$^\texttt{*}$ & no & 4.70042 & 1.74281 & 14.0868 & 7.00601 & 9.36609 & 3.24191 & 21.3807 & 10.3596 \\
char 20-gram$^\texttt{*}$ & no & 4.50257 & 1.67847 & 13.9288 & 6.84336 & 9.40996 & 3.18202 & 21.6887 & 10.2693 \\
\midrule
word ConvLM$^\texttt{*}$ & yes &  2.74486 & 1.04841 & 10.0888 & 5.21076 &  9.93639 & 3.52138 & 19.3876 & 9.40626 \\
char ConvLM-20B & yes & 3.09057 & 1.20086 & 10.8555 & 5.64639 & 10.3751 & 3.4056 & 18.844 & 9.55678 \\
char ConvLM-20B & no & 3.13847 & 1.21178 & 10.9581 & 5.6851 & 7.74293 & 2.65501 & 18.3367 & 9.37615 \\
\bottomrule
\end{tabular}
\end{table}

\paragraph{Beam size analysis}
For each utterance we can define an effective beam size as the maximum position index in the sorted beam over all frames for the final transcription. In other words, selecting a beam size larger than the effective beam size should not affect decoding. In Figure~\ref{Fig:beam_pdf}, we show that the effective beam size for the character-level models is significantly smaller than one for the word-level models. When decoding with word-level LMs, a large (2000-8000) beam size is often needed, while for the character-level LMs, a beam size of 2000 is always enough.
This property looks promising from the computational point of view, when switching to character-level LMs.

\paragraph{Lexicon-free decoding analysis}
We investigated how out-of-vocabulary (OOV) words are transcribed by our lexicon-free decoder, as those words cannot be output by a standard beam-search decoder conditioned with a word-level lexicon.
The OOV words we consider are not present in the lexicon vocabulary, meaning these words are beyond the top-$k$ most frequent words chosen in our lexicon, or did not even appear in the acoustic and language models training/validation sets.
In Table~\ref{T:Decoder:Librispeech_unk}, we evaluated WER and CER on the isolated ``out-of-vocabulary'' utterances, which contain at least one OOV word. For comparison, we also report performance on ``in vocabulary'' utterances (IV), which contain only words present in the lexicon. 
The lexicon-free decoder performs significantly better on the utterances that include OOV words, while holding competitive performance on the IV utterances.
The lexicon-free decoder with an $n$-gram language model recognizes up to 25\% (28.5\%) and 13.5\% (13.5\%) OOV words (occurrences) for the {\it clean} and {\it other} test parts of Librispeech, respectively. With a ConvLM language model, this performance raises up to 31\% (33\%) and 10\% (8\%) OOV words (occurrences) for the {\it clean} and {\it other} test parts of Librispeech, respectively. 
A few examples of decoded transcriptions are reported in Table~\ref{T:LM:examples}.

\begin{table}[t!]
\caption{Librispeech transcriptions for decoder with $n$-gram LMs: target (T), word-level LM and lexicon (W), character-level LM and lexicon (C), lexicon-free decoder with character-level LM (F). Underlined words are not presented in the lexicon.}
\label{T:LM:examples}
\centering
\setlength\tabcolsep{2pt}
\begin{tabular}{r l}
\toprule
T & {\bf\underline{fauchelevent}} limped along behind the hearse in a very contented frame of mind \\
W & {\bf lochleven} limped along behind the hearse in a very contented frame of mind \\
C & {\bf lochleven} limped along behind the hearse in a very contented frame of mind \\
F & {\bf \underline{fauchelevent}} limped along behind the hearse in a very contented frame of mind\\ 
\midrule
T & ... he did not want to join his own friends that is sergey ivanovitch stepan {\bf \underline{arkadyevitch} \underline{sviazhsky}} and ... \\
W & ... he did not come to join his own friends that a soldier ivanovitch step on {\bf markovitch the sky} and ... \\
C & ... he did not own to join his own friends that a sojer ivanovitch step on {\bf radovitch totski} and ... \\
F & ... he did not own to join his own friends that a sojer ivanovitch stepan {\bf \underline{arkadyevitch} \underline{tievski}} and ... \\
\midrule
T & menahem king of israel had died and was succeeded by his son {\bf \underline{pekahiah}} \\
W & many a king of israel had died and was succeeded by his son {\bf pekah} \\
C & many king of israel had died and was succeeded by his son {\bf pekah} \\
F & many a king of israel had died and was succeeded by his son {\bf \underline{pekaiah}} \\ 
\bottomrule
\end{tabular}
\end{table}

\section{Conclusion}
We built an ASR system with a beam-search decoder based on a character-level language model that achieves performance close to the state-of-the-art among end-to-end models.
We also showed that transcribing with a lexicon-free beam-search decoder achieves a performance similar to transcribing with a beam-search decoder conditioned by a word-level lexicon, itself already close to the state-of-the-art.
Moreover, lexicon-free ASR naturally handles out-of-vocabulary words: it has significantly better performance on the utterances with out-of-vocabulary words than when the system is conditioned by a word-level lexicon.

\section{Acknowledgements}
We thank Qiantong Xu for helpful discussions.

\section{Addendum}
In the Interspeech2019 submission paper lower bounds on the word perplexity for character-level language models were computed excluding end of sentence token while estimating the denominator from (\ref{form:denom}). In Tables~\ref{T:LM:WSJ} and~\ref{T:LM:Librispeech} updated results are presented where the end of sentence is correctly accounted. This does not affect any outcomes and conclusions in the paper.

The code for experiments is open-sourced and available at \url{https://github.com/facebookresearch/wav2letter/tree/master/recipes/models/lexicon_free}.

\bibliographystyle{unsrt}  

\end{document}